%% file: paper.tex
\documentclass[11pt]{article}
\usepackage[utf8]{inputenc}
\usepackage[T1]{fontenc}
\usepackage{graphicx}
\usepackage{grffile}
\usepackage{longtable}
\usepackage{wrapfig}
\usepackage{rotating}
\usepackage[normalem]{ulem}
\usepackage{amsmath}
\usepackage{textcomp}
\usepackage{amssymb}
\usepackage{capt-of}
\usepackage{hyperref}
\usepackage[nonatbib,final]{nips_2017}
\usepackage[utf8]{inputenc} 
\usepackage[T1]{fontenc}    
\usepackage{textcomp}
\usepackage{hyperref}       
\usepackage{url}            
\usepackage{booktabs}       
\usepackage{multirow}
\usepackage[table,svgnames]{xcolor}
\usepackage{amsfonts}       
\usepackage{nicefrac}       
\usepackage{microtype}      
\usepackage{cleveref}
\usepackage{tikz}
\usepackage{bm}
\usepackage{xspace}
\usepackage{subcaption}
\usepackage{enumitem}
\usepackage[numbers]{natbib}
\renewcommand{\vec}[1]{\bm{#1}}

\newcommand{\vu}{\vec{u}}
\newcommand{\vv}{\vec{v}}
\newcommand{\vr}{\vec{r}}
\newcommand{\vt}{\vec{\theta}}
\newcommand{\vx}{\vec{x}}

\newcommand{\lossfn}{\mathcal{L}}
\newcommand{\Set}[1]{\mathcal{#1}}
\newcommand{\manifold}[1]{\mathcal{#1}}
\newcommand{\acosh}{\operatorname{arcosh}}
\newcommand{\dist}{d}
\newcommand{\tansp}{\mathcal{T}}
\newcommand{\hsp}{\mathbb{H}}
\newcommand{\tanspt}{\tansp_{\theta} \manifold{B}}

\newcommand{\R}{\mathbb{R}}

\newcommand{\cellhi}{\cellcolor{blue!10}}
\newcommand{\cellbest}{\bf}
\DeclareMathOperator*{\argmin}{arg\,min}
\newcommand{\method}[1]{\textsc{#1}\xspace}
\newcommand{\score}{\operatorname{score}}
\newcommand{\retract}{\mathfrak{R}}
\setlist[description]{leftmargin=10pt}
\newcolumntype{H}{>{\setbox0=\hbox\bgroup}c<{\egroup}@{}}
\author{Maximilian Nickel\\
Facebook AI Research\\
\texttt{maxn@fb.com}
\And
Douwe Kiela\\
Facebook AI Research\\
\texttt{dkiela@fb.com}}
\makeatletter
\renewcommand{\@noticestring}{}
\makeatother
\date{\today}
\title{Poincaré Embeddings for \\  Learning Hierarchical Representations}
\hypersetup{
 pdfauthor={Maximilian Nickel <maxn@fb.com>},
 pdftitle={Poincaré Embeddings for \\  Learning Hierarchical Representations},
 pdfkeywords={},
 pdfsubject={},
 pdfcreator={Emacs 25.2.1 (Org mode 9.0.6)}, 
 pdflang={English}}
\begin{document}

\maketitle
\begin{abstract}
Representation learning has become an invaluable approach for learning from
symbolic data such as text and graphs. However, while complex symbolic datasets
often exhibit a latent hierarchical structure, state-of-the-art methods typically
learn embeddings in Euclidean vector spaces, which do not account for this property. 
For this purpose, we introduce a new approach for
learning hierarchical representations of symbolic data by embedding them into
hyperbolic space -- or more precisely into an \(n\)-dimensional Poincaré ball.
Due to the underlying hyperbolic geometry, this allows us to learn parsimonious
representations of symbolic data by simultaneously capturing hierarchy and
similarity. We introduce an efficient algorithm to learn the embeddings based on
Riemannian optimization and show experimentally that Poincaré embeddings
outperform Euclidean embeddings significantly on data with latent hierarchies,
both in terms of representation capacity and in terms of generalization ability.
\end{abstract}

\section{Introduction}
\label{sec:intro}
Learning representations of symbolic data such as text, graphs and
multi-relational data has become a central paradigm in machine learning and
artificial intelligence. For instance, word embeddings such as \method{word2vec}
\cite{DBLP:journals/corr/MikolovSCCD13}, \method{GloVe} \cite{pennington2014glove}
and \method{FastText} \cite{bojanowski2016enriching} are widely used for tasks
ranging from machine translation to sentiment analysis. Similarly, embeddings of
graphs such as latent space embeddings \cite{hoff2002latent}, \method{node2vec}
\cite{grover2016node2vec}, and \method{DeepWalk} \cite{perozzi2014deepwalk} have
found important applications for community detection and link prediction in
social networks. Embeddings of multi-relational data such as \method{Rescal}
\cite{DBLP:conf/icml/NickelTK11}, \method{TransE}
\cite{DBLP:conf/nips/BordesUGWY13}, and Universal Schema \cite{riedel2013relation}
are being used for knowledge graph completion and information extraction.

Typically, the objective of embedding methods is to organize symbolic objects
(e.g., words, entities, concepts) in a way such that their similarity in the
embedding space reflects their semantic or functional similarity. For this
purpose, the similarity of objects is usually measured either by their distance
or by their inner product in the embedding space. For instance,
\citet{DBLP:journals/corr/MikolovSCCD13} embed words in \(\R^d\) such that their
inner product is maximized when words co-occur within similar contexts in text
corpora. This is motivated by the distributional hypothesis
\cite{harris1954distributional,firth1957synopsis}, i.e., that the meaning of words
can be derived from the contexts in which they appear. Similarly,
\citet{hoff2002latent} embed social networks such that the
distance between social actors is minimized if they are connected in the
network. This reflects the homophily property found in many real-world
networks, i.e. that similar actors tend to associate with each other.

Although embedding methods have proven successful in numerous applications, they
suffer from a fundamental limitation: their ability to model complex patterns is
inherently bounded by the dimensionality of the embedding space. For instance,
\citet{DBLP:conf/nips/NickelJT14} showed that linear embeddings of graphs can
require a prohibitively large dimensionality to model certain types of
relations. Although non-linear embeddings can mitigate this problem
\cite{bouchard2015approximate}, complex graph patterns can still
require a computationally infeasible embedding dimensionality. As a consequence, no
method yet exists that is able to compute embeddings of large graph-structured
data -- such as social networks, knowledge graphs or taxonomies -- without loss of
information. Since the ability to express information is a precondition for
learning and generalization, it is therefore important to increase the
representation capacity of embedding methods such that they can realistically be
used to model complex patterns on a large scale.
In this work, we focus on mitigating this problem for
a certain class of symbolic data, i.e., large datasets whose objects can be
organized according to a latent hierarchy -- a property that is inherent in
many complex datasets. For instance, the existence of power-law distributions in
datasets can often be traced back to hierarchical structures
\cite{ravasz2003hierarchical}. Prominent examples of power-law distributed data
include natural language (Zipf's law \cite{DBLP:books/aw/Zipf49}) and scale-free
networks such as social and semantic networks \cite{steyvers2005large}. Similarly,
the empirical analysis of \citet{adcock2013tree} indicated that many real-world
networks exhibit an underlying tree-like structure.

To exploit this structural property for learning more efficient representations,
we propose to compute embeddings not in Euclidean but in hyperbolic space, i.e.,
space with constant negative curvature. Informally, hyperbolic space can be
thought of as a continuous version of trees and as such it is naturally equipped
to model hierarchical structures. For instance, it has been shown that any
finite tree can be embedded into a finite hyperbolic space such that distances
are preserved approximately \cite{gromov1987hyperbolic}. We base our approach on a
particular model of hyperbolic space, i.e., the Poincaré ball model, as it is
well-suited for gradient-based optimization. This allows us to develop an
efficient algorithm for computing the embeddings based on Riemannian
optimization, which is easily parallelizable and scales to large datasets.
Experimentally, we show that our approach can provide high quality embeddings of
large taxonomies -- both with and without missing data. Moreover, we show that
embeddings trained on \method{WordNet} provide state-of-the-art performance for
lexical entailment. On collaboration networks, we also show that Poincaré
embeddings are successful in predicting links in graphs where they outperform
Euclidean embeddings, especially in low dimensions.

The remainder of this paper is organized as follows: In \Cref{sec:hyperbolic} we
briefly review hyperbolic geometry and discuss related work regarding hyperbolic
embeddings. In \Cref{sec:poincare} we introduce Poincaré embeddings and discuss
how to compute them. In \Cref{sec:experiments} we evaluate our approach on tasks
such as taxonomy embedding, link prediction in networks and predicting lexical entailment.

\section{Embeddings and Hyperbolic Geometry}
\label{sec:hyperbolic}
\begin{figure}
  \centering
  \begin{minipage}[b]{.3\linewidth}
    \centering
    \resizebox{\linewidth}{!}{\input{geodesics.pgf}}
    \subcaption{Geodesics of the Poincaré disk}\label{fig:geodesics}
  \end{minipage}
  \hfill
  \begin{minipage}[b]{.3\linewidth}
    \centering
    \includegraphics[width=.9\linewidth,clip,trim={0cm 0cm 0cm 0cm}]{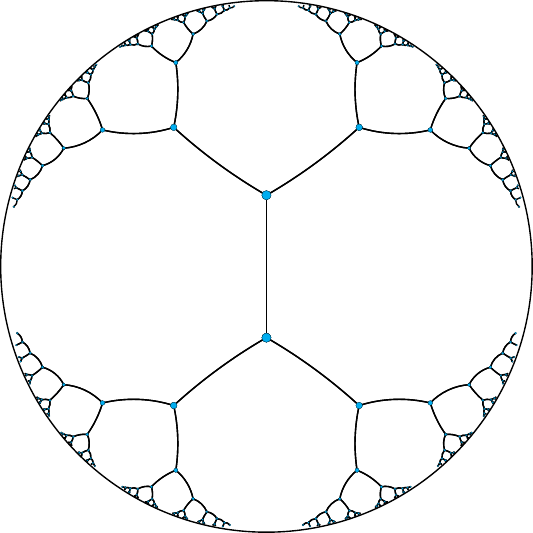}
    \subcaption{Embedding of a tree in $\manifold{B}^2$}\label{fig:tree-embedding-example}
  \end{minipage}
  \hfill
  \begin{minipage}[b]{.3\linewidth}
    \centering
    \includegraphics[width=\linewidth,clip,trim={2cm 0cm 1cm 2cm}]{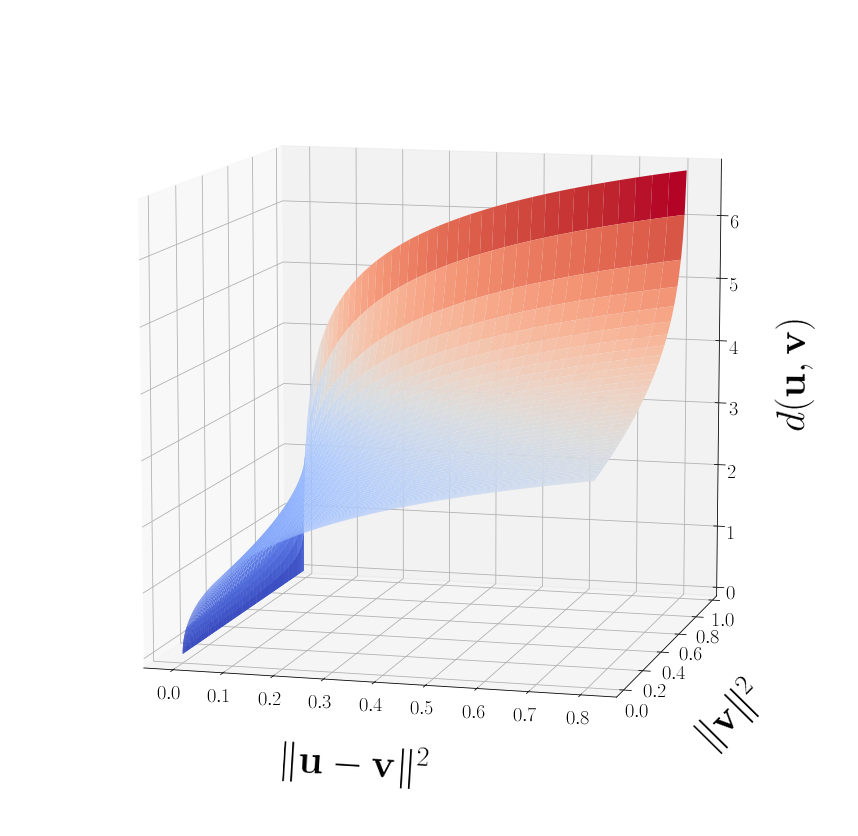}
    \subcaption{Growth of Poincaré distance}\label{fig:poin-dist}
  \end{minipage}
  \caption{\small\subref{fig:geodesics} Due to the negative curvature of $\manifold{B}$, the distance of points increases exponentially (relative to their Euclidean distance) the closer they are to the boundary. \subref{fig:poin-dist} Growth of the Poincaré distance $d(\vu,\vv)$ relative to the Euclidean distance and the norm of $\vv$ (for fixed $\|\vu\| = 0.9$). \subref{fig:tree-embedding-example} Embedding of a regular tree in $\manifold{B}^2$ such that all connected nodes are spaced equally far apart (i.e., all black line segments have identical hyperbolic length).}
\end{figure}

Hyperbolic geometry is a non-Euclidean geometry which studies spaces of constant
negative curvature. It is, for instance, associated with Minkowski spacetime in
special relativity. In network science, hyperbolic spaces have started to
receive attention as they are well-suited to model hierarchical data. For
instance, consider the task of embedding a tree into a metric space such that
its structure is reflected in the embedding. A regular tree with branching
factor \(b\) has \((b + 1)b^{\ell-1}\) nodes at level \(\ell\) and \(((b + 1)b^\ell - 2) /
(b - 1)\) nodes on a level less or equal than \(\ell\). Hence, the number of children
grows exponentially with their distance to the root of the tree. In hyperbolic
geometry this kind of tree structure can be modeled easily in two dimensions:
nodes that are \emph{exactly} \(\ell\) levels below the root are placed on a sphere in
hyperbolic space with radius \(r \propto \ell\) and nodes that are \emph{less than} \(\ell\) levels
below the root are located within this sphere. This type of construction is
possible as hyperbolic disc area and circle length grow exponentially with their
radius.\footnote{For instance, in a two dimensional hyperbolic space with constant
curvature \(K=-1\), the length of a circle is given as \(2 \pi \sinh r\) while the
area of a disc is given as \(2\pi (\cosh r - 1)\). Since \(\sinh r =
\frac{1}{2}(e^r - e^{-r})\) and \(\cosh r = \frac{1}{2}(e^r + e^{-r})\), both disc
area and circle length grow exponentially with \(r\).} See
\Cref{fig:tree-embedding-example} for an example. Intuitively, hyperbolic spaces
can be thought of as continuous versions of trees or vice versa, trees can be
thought of as "discrete hyperbolic spaces"
\cite{hyperbolic/krioukov2010hyperbolic}. In \(\R^2\), a similar construction is not
possible as circle length (\(2\pi r\)) and disc area (\(2\pi r^2\)) grow only linearly
and quadratically with regard to \(r\) in Euclidean geometry. Instead, it is
necessary to increase the dimensionality of the embedding to model increasingly
complex hierarchies. As the number of parameters increases, this can lead to
computational problems in terms of runtime and memory complexity as well as to
overfitting.

Due to these properties, hyperbolic space has recently been considered to model
complex networks. For instance, \citet{kleinberg2007geographic} introduced
hyperbolic geometry for greedy routing in geographic communication networks.
Similarly, \citet{boguna2010sustaining} proposed hyperbolic embeddings of the AS
Internet topology to perform greedy shortest path routing in the embedding
space. \citet{hyperbolic/krioukov2010hyperbolic} developed a framework to model
complex networks using hyperbolic spaces and discussed how typical properties
such as heterogeneous degree distributions and strong clustering emerges by
assuming an underlying hyperbolic geometry to these networks.
\citet{adcock2013tree} proposed a measure based on Gromov's \(\delta\)-hyperbolicity
\cite{gromov1987hyperbolic} to characterize the tree-likeness of graphs.

In machine learning and artificial intelligence on the other hand, Euclidean
embeddings have become a popular approach for learning from symbolic data. For
instance, in addition to the methods discussed in \Cref{sec:intro},
\citet{paccanaro2001learning} proposed one of the first embedding methods to learn
from relational data. More recently, Holographic \cite{kgs/nickel2016holographic}
and Complex Embeddings \cite{DBLP:conf/icml/TrouillonWRGB16} have shown
state-of-the-art performance in Knowledge Graph completion. In relation to
hierarchical representations, \citet{vilnis2015word} proposed to learn density-based
word representations, i.e., Gaussian embeddings, to capture uncertainty
and asymmetry. Given information about hierarchical relations in the form of
ordered input pairs, \citet{vendrov2015order} proposed Order Embeddings to model
visual-semantic hierarchies over words, sentences, and images.

\section{Poincaré Embeddings}
\label{sec:poincare}
In the following, we are interested in finding embeddings of symbolic data such
that their distance in the embedding space reflects their semantic similarity.
We assume that there exists a latent hierarchy in which the symbols can be
organized. In addition to the similarity of objects, we intend to also reflect
this hierarchy in the embedding space to improve over existing methods in two
ways:
\begin{enumerate}
\item By inducing an appropriate bias on the structure of the embedding space, we
aim at learning more parsimonious embeddings for superior generalization
performance and decreased runtime and memory complexity.
\item By capturing the hierarchy explicitly in the embedding space, we aim at
gaining additional insights about the relationships between symbols and the
importance of individual symbols.
\end{enumerate}

However, we do not assume that we have direct access to information about the
hierarchy, e.g., via ordered input pairs. Instead, we consider the task of
inferring the hierarchical relationships fully unsupervised, as is, for
instance, necessary for text and network data. For these reasons -- and
motivated by the discussion in \Cref{sec:hyperbolic} -- we embed symbolic data
into hyperbolic space \(\hsp\). In contrast to Euclidean space \(\R\), there exist
multiple, equivalent models of \(\hsp\) such as the Beltrami-Klein model, the
hyperboloid model, and the Poincaré half-plane model. In the following, we will
base our approach on the Poincaré ball model, as it is well-suited for
gradient-based optimization.\footnote{It can be seen easily that the distance
function of the Poincare ball in \Cref{eq:distance} is differentiable. Hence,
for this model, an optimization algorithm only needs to maintain the constraint
that \(\|\vx\| < 1\) for all embeddings. Other models of hyperbolic space however,
would be more more difficult to optimize, either due to the form of their
distance function or due to the constraints that they introduce. For instance,
the hyperboloid model is constrained to points where \(\langle \vx, \vx \rangle = -1\), while
the distance function of the Beltrami-Klein model requires to compute the
location of ideal points on the boundary \(\partial \manifold{B}\) of the unit ball.} In
particular, let \({\manifold{B}^d = \{\vx \in \R^d\ |\ \|\vx\| < 1\}}\) be the \emph{open}
\(d\)-dimensional unit ball, where \(\|\cdot\|\) denotes the Euclidean norm. The
Poincaré ball model of hyperbolic space corresponds then to the Riemannian
manifold \((\manifold{B}^d, g_{\vx})\), i.e., the open unit ball equipped with the
Riemannian metric tensor
\begin{equation*}
  g_{\vx} = \left( \frac{2}{1 - \|\vx\|^2} \right)^2 g^E ,
\end{equation*}
where \(\vx \in \manifold{B}^d\) and \(g^E\) denotes the Euclidean metric tensor.
Furthermore, the distance between points \(\vu, \vv \in \manifold{B}^d\) is given as
\begin{align}
  \dist(\vu, \vv) & = \acosh\left(1 + 2\frac{\|\vu - \vv\|^2}{(1 - \|\vu\|^2)(1 - \|\vv\|^2)}\right) \label{eq:distance} .
\end{align}
The boundary of the ball is denoted by \(\partial\manifold{B}\). It corresponds to the
sphere \(\manifold{S}^{d-1}\) and is not part of the hyperbolic space, but represents
infinitely distant points. Geodesics in \(\manifold{B}^d\) are then circles that are
orthogonal to \(\partial \manifold{B}\) (as well as all diameters). See
\Cref{fig:geodesics} for an illustration.

It can be seen from \Cref{eq:distance}, that the distance within the Poincaré
ball changes smoothly with respect to the location of \(\vu\) and \(\vv\). This
locality property of the Poincaré distance is key for finding continuous
embeddings of hierarchies. For instance, by placing the root node of a tree at
the origin of \(\manifold{B}^d\) it would have a relatively small distance to all
other nodes as its Euclidean norm is zero. On the other hand, leaf nodes can be
placed close to the boundary of the Poincaré ball as the distance grows very
fast between points with a norm close to one. Furthermore, please note that
\Cref{eq:distance} is symmetric and that the hierarchical organization of the
space is solely determined by the distance of points to the origin. Due to this
self-organizing property, \Cref{eq:distance} is applicable in an unsupervised
setting where the hierarchical order of objects is not specified in advance such
as text and networks. Remarkably, \Cref{eq:distance} allows us therefore to
learn embeddings that simultaneously capture the hierarchy of objects (through
their norm) as well a their similarity (through their distance).

Since a single hierarchical structure can already be represented in two
dimensions, the Poincaré disk (\(\manifold{B}^2\)) is typically used to represent
hyperbolic geometry. In our method, we instead use the Poincaré ball
(\(\manifold{B}^d\)), for two main reasons: First, in many datasets such as text
corpora, multiple latent hierarchies can co-exist, which can not always be
modeled in two dimensions. Second, a larger embedding dimension can decrease the
difficulty for an optimization method to find a good embedding (also for single
hierarchies) as it allows for more degrees of freedom during the optimization
process.

To compute Poincaré embeddings for a set of symbols \(\Set{S} = \{x_i\}_{i=1}^n\), we
are then interested in finding embeddings \(\Theta = \{\vt_i\}_{i=1}^n\), where \(\vt_i \in
\manifold{B}^d\). We assume we are given a problem-specific loss function
\(\lossfn(\Theta)\) which encourages semantically similar objects to be close in the
embedding space according to their Poincaré distance. To estimate \(\Theta\), we then
solve the optimization problem
\begin{equation}
\Theta^\prime \gets \argmin_{\Theta} \lossfn(\Theta) \quad\quad \text{s.t. } \forall\, \vt_i \in \Theta: \|\vt_i\| < 1 .\label{eq:loss}
\end{equation}
We will discuss specific loss functions in \Cref{sec:experiments}.

\subsection{Optimization}
\label{sec:optim}
Since the Poincaré Ball has a Riemannian manifold structure, we can optimize
\Cref{eq:loss} via stochastic Riemannian optimization methods such as RSGD
\cite{optimization/bonnabel2013stochastic} or RSVRG
\cite{optimization/zhang2016Riemannian}. In particular, let \(\tanspt\) denote the
tangent space of a point \(\vt \in \manifold{B}^d\). Furthermore, let \({\nabla_R \in
\tanspt}\) denote the Riemannian gradient of \(\lossfn(\vt)\) and let \(\nabla_E\)
denote the Euclidean gradient of \(\lossfn(\vt)\). Using RSGD, parameter
updates to minimize \Cref{eq:loss} are then of the form
\begin{equation*}
  \vt_{t+1} = \retract_{\theta_t}\left(-\eta_t \nabla_R \lossfn(\vt_t) \right)
\end{equation*}
where \(\retract_{\theta_t}\) denotes the retraction onto \(\manifold{B}\) at \(\vt\) and
\(\eta_t\) denotes the learning rate at time \(t\). Hence, for the minimization of
\Cref{eq:loss}, we require the Riemannian gradient and a suitable retraction.
Since the Poincaré ball is a conformal model of hyperbolic space, the angles
between adjacent vectors are identical to their angles in the Euclidean space.
The length of vectors however might differ. To derive the Riemannian gradient
from the Euclidean gradient, it is sufficient to rescale \(\nabla_E\) with the inverse
of the Poincaré ball metric tensor, i.e., \(g^{-1}_\theta\). Since \(g_\theta\) is a
scalar matrix, the inverse is trivial to compute. Furthermore, since
\Cref{eq:distance} is fully differentiable, the Euclidean gradient can easily be
derived using standard calculus. In particular, the Euclidean gradient \(\nabla_E =
\frac{\partial \lossfn(\vt)}{\partial d(\vt, \vx)} \frac{\partial d(\vt, \vx)}{\partial \vt}\) depends on the
gradient of \(\lossfn\), which we assume is known, and the partial derivatives of
the Poincaré distance, which can be computed as follows: Let \(\alpha = 1 - \|\vt\|^2\)
, \(\beta = 1 - \|\vx\|^2\) and let
\begin{equation}
  \gamma = 1 + \frac{2}{\alpha \beta} \|\vt -\vx\|^2
\end{equation}
The partial derivate of the Poincaré distance with respect to \(\vt\) is then given as
\begin{equation}
  \frac{\partial d(\vt, \vx)}{\partial \vt} = \frac{4}{\beta \sqrt{\gamma^2 - 1}} \left(\frac{\|\vx\|^2 - 2\langle\vt, \vx\rangle + 1}{\alpha^2} \vt - \frac{\vx}{\alpha} \right) . \label{eq:partial}
\end{equation}
Since \(d(\cdot, \cdot)\) is symmetric, the partial derivative \(\frac{\partial d(\vx, \vt)}{\partial
\vt}\) can be derived analogously. As retraction operation we use
\(\retract_\theta(\vv) = \vt + \vv\). In combination with the Riemannian gradient,
this corresponds then to the well-known natural gradient method
\cite{DBLP:journals/neco/Amari98}. Furthermore, we constrain the embeddings to
remain within the Poincaré ball via the projection
\begin{equation*}
  \text{proj}(\vt) = \begin{cases}
  \vt / \|\vt\| - \varepsilon & \text{if }\|\vt\| \geq 1 \\
  \vt & \text{otherwise ,}
  \end{cases}
\end{equation*}
where \(\varepsilon\) is a small constant to ensure numerical stability. In all experiments we used \(\varepsilon = 10^{-5}\).
In summary, the full update for a single embedding is then of the form
\begin{equation}
  \vt_{t+1} \gets \text{proj}\left(\vt_t - \eta_t \frac{(1 - \|\vt_t\|^2)^2}{4} \nabla_E \right) \label{eq:update} .
\end{equation}

It can be seen from \Cref{eq:partial,eq:update} that this algorithm scales well
to large datasets, as the computational and memory complexity of an update depends linearly
on the embedding dimension. Moreover, the algorithm is straightforward to parallelize via
methods such as Hogwild \cite{DBLP:conf/nips/RechtRWN11}, as the updates are
sparse (only a small number of embeddings are modified in an update) and
collisions are very unlikely on large-scale data.

\subsection{Training Details}
\label{sec:org5b48fd6}
In addition to this optimization procedure, we found that the following training
details were helpful for obtaining good representations: First, we initialize
all embeddings randomly from the uniform distribution \({\Set{U}(-0.001,0.001)}\).
This causes embeddings to be initialized close to the origin of \(\manifold{B}^d\).
Second, we found that a good initial angular layout can be helpful to find good
embeddings. For this reason, we train during an initial "burn-in" phase with a
reduced learning rate \(\eta / c\). In combination with initializing close to the
origin, this can improve the angular layout without moving too far towards the
boundary. In our experiments, we set \({c=10}\) and the duration of the burn-in to
10 epochs.

\section{Evaluation}
\label{sec:experiments}
In this section, we evaluate the quality of Poincaré embeddings for a variety of
tasks, i.e., for the embedding of taxonomies, for link prediction in networks,
and for modeling lexical entailment. We compare the \textbf{Poincaré} distance
as defined in \Cref{eq:distance} to the following two distance functions:
\begin{description}
\item[{Euclidean}] In all cases, we include the Euclidean distance \({d(\vu,\vv) =
               \|\vu - \vv\|^2}\). As the Euclidean distance is flat and
symmetric, we expect that it requires a large dimensionality
to model the hierarchical structure of the data.
\item[{Translational}] For asymmetric data, we also include the score
function \({d(\vu, \vv) = \|\vu - \vv + \vr\|^2}\), as proposed
by \citet{DBLP:conf/nips/BordesUGWY13} for modeling large-scale
graph-structured data. For this score function, we also learn
the global translation vector \(\vr\) during training.
\end{description}
Note that the translational score function has, due to its asymmetry, more
information about the nature of an embedding problem than a symmetric distance
when the order of \((u,v)\) indicates the hierarchy of elements. This is, for
instance, the case for \({\texttt{is-a}(u,v)}\) relations in taxonomies. For the
Poincaré distance and the Euclidean distance we could randomly permute the order
of \((u,v)\) and obtain the identical embedding, while this is not the case for
the translational score function. As such, it is not fully unsupervised and
only applicable where this hierarchical information is available.

\subsection{Embedding Taxonomies}
\label{sec:taxonomies}
In the first set of experiments, we are interested in evaluating the ability of
Poincaré embeddings to embed data that exhibits a clear latent hierarchical structure. 
For this purpose, we conduct experiments on the \emph{transitive closure} of the \method{WordNet}
noun hierarchy \cite{miller1998wordnet} in two settings:

\begin{description}
\item[{Reconstruction}] To evaluate representation capacity, we embed fully observed data and
reconstruct it from the embedding. The reconstruction error
in relation to the embedding dimension is then a measure for
the capacity of the model.
\item[{Link Prediction}] To test generalization performance, we split the data
into a train, validation and test set by randomly holding out observed
links. Links in the validation and test set do not include the root or
leaf nodes as these links would either be trivial to predict or impossible
to predict reliably.
\end{description}

Since we are using the transitive closure, the hypernymy relations form a directed acyclic graph 
such that the hierarchical structure is not directly visible from the raw data but has to be inferred. The
transitive closure of the \method{WordNet} noun hierarchy consists of 82,115
nouns and 743,241 hypernymy relations. On this data, we learn embeddings
in both settings as follows: Let \(\Set{D} = \{(u, v)\}\) be the set of observed
hypernymy relations between noun pairs. We then learn embeddings of all symbols
in \(\Set{D}\) such that related objects are close in the embedding space. In
particular, we minimize the loss function
\begin{equation}
  \lossfn(\Theta) = \sum_{(u,v) \in \Set{D}}\log \frac{e^{-d(\vu,\vv)}}{\sum_{\vv^\prime \in \Set{N}(u)} e^{-d(\vu,\vv^\prime)}} , \label{eq:soft-ranking}
\end{equation}
where \(\Set{N}(u) = \{v\ |\ (u, v) \not \in \Set{D}\} \cup \{u\}\) is the set of negative
examples for \(u\) (including \(u\)). For training, we randomly sample 10 negative
examples per positive example. \Cref{eq:soft-ranking} can be interpreted as a
soft ranking loss where related objects should be closer than objects for which
we didn't observe a relationship. This choice of loss function is motivated by
the fact that we don't want to push symbols belonging to distinct
subtrees arbitrarily far apart as their subtrees might still be close. Instead
we want them to be farther apart than symbols with an observed relation.

\begin{table}
  \centering
  \small
  \caption{Experimental results on the transitive closure of the \method{WordNet} noun hierarchy. Highlighted cells indicate the best Euclidean embeddings as well as the Poincaré embeddings which acheive equal or better results. Bold numbers indicate absolute best results.\label{tab:embedding-nouns}}
  \begin{tabular}{lllHcccccc}
    \toprule
    & & & \multicolumn{7}{c}{\textbf{Dimensionality}} \\
    \cmidrule(l){4-10}
    & & &  2 &  5  & 10  & 20  & 50  & 100 & 200 \\
    \midrule
    \multirow{7}{*}{\rotatebox[origin=c]{90}{\shortstack{\method{WordNet} \\ \textbf{Reconstruction}}}} & 
    \multirow{2}{*}{\textbf{Euclidean}} 
    & Rank & 9085.6 & 3542.3 & 2286.9 & 1685.9 & 1281.7 & 1187.3 & \cellhi 1157.3 \\
    & & MAP & 0.003 & 0.024 & 0.059 & 0.087 & 0.140 & 0.162 & \cellhi 0.168 \\
    \cmidrule(l){2-10}
    & \multirow{2}{*}{\textbf{Translational}} 
    & Rank & 383.9 & 205.9 & 179.4 & 95.3 & 92.8 & 92.7 & \cellhi 91.0 \\
    & & MAP & 0.426 & 0.517 & 0.503 & 0.563 & 0.566 & 0.562 & \cellhi 0.565 \\
    \cmidrule(l){2-10}
    & \multirow{2}{*}{\textbf{Poincaré}} 
    & Rank & 75.1 & \cellhi 4.9 & 4.02 & 3.84 & 3.98 & 3.9 & \cellbest 3.83 \\
    & & MAP & - & \cellhi 0.823 & 0.851 & 0.855  & 0.86 & 0.857 & \cellbest 0.87 \\
    \midrule
    \multirow{7}{*}{\rotatebox[origin=c]{90}{\shortstack{\method{WordNet} \\ \textbf{Link Pred.}}}} & 
    \multirow{2}{*}{\textbf{Euclidean}} 
    & Rank & 7646.2 & 3311.1 & 2199.5 & 952.3 & 351.4 & 190.7 & \cellhi 81.5 \\
    & & MAP & 0.003 & 0.024 & 0.059 & 0.176 & 0.286 & 0.428 & \cellhi 0.490 \\
    \cmidrule(l){2-10}
    & \multirow{2}{*}{\textbf{Translational}} 
    & Rank & 133.7 & 65.7 & 56.6 & 52.1 & 47.2 & 43.2 & \cellhi 40.4 \\
    & & MAP & 0.484 & 0.545 & 0.554 & 0.554 & 0.56 & 0.562 & \cellhi 0.559 \\
    \cmidrule(l){2-10}
    & \multirow{2}{*}{\textbf{Poincaré}} 
    & Rank & 76.6 & \cellhi 5.7 & \cellbest 4.3 & 4.9 & 4.6 & 4.6 & 4.6 \\
    & & MAP & - & \cellhi 0.825 & 0.852 & 0.861 & \cellbest 0.863 & 0.856 & 0.855 \\
    \bottomrule
  \end{tabular}
\end{table}

\begin{figure}
  \centering
  \begin{minipage}[b]{0.45\linewidth}
    \includegraphics[width=0.9\linewidth]{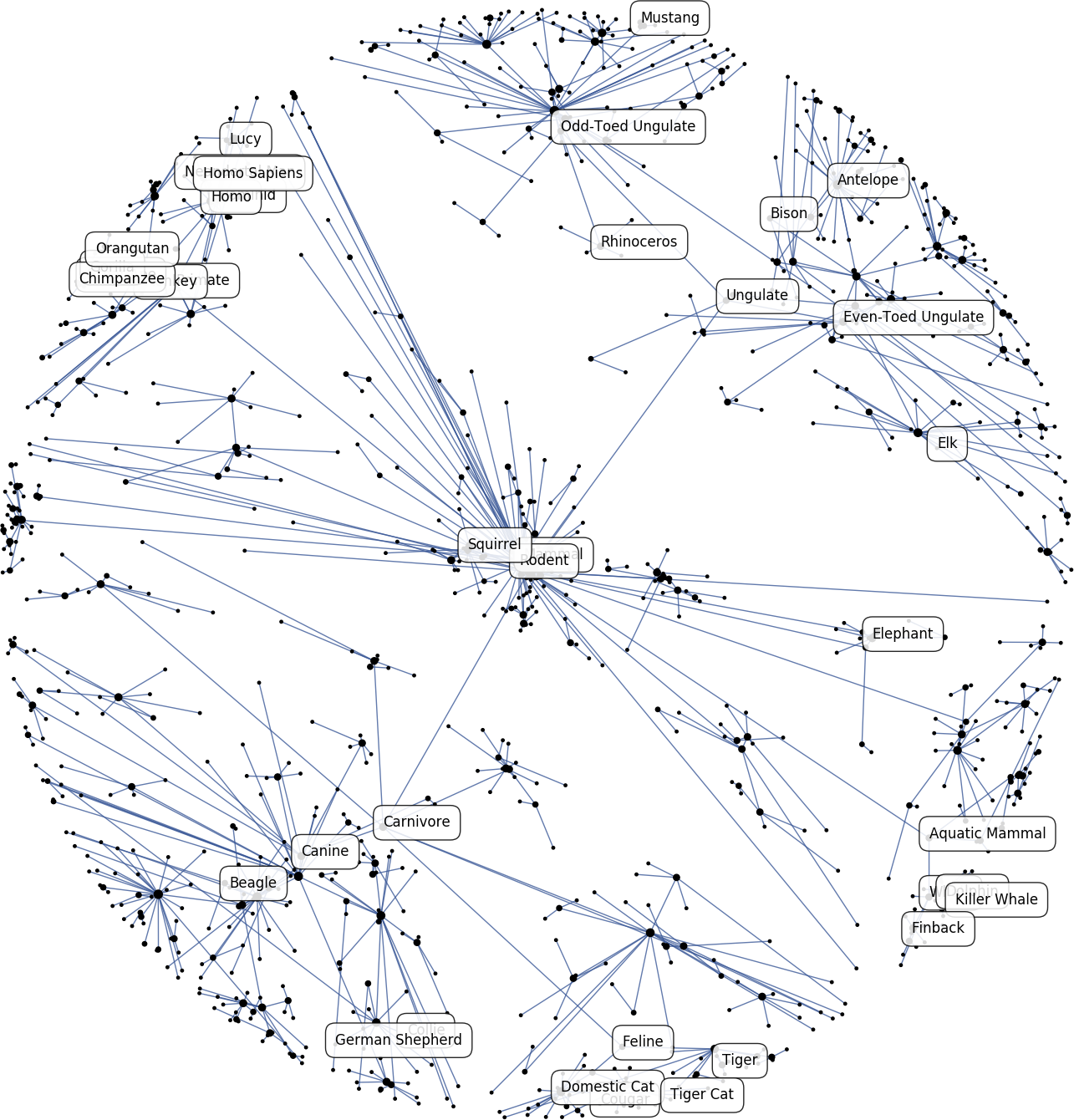}
    \subcaption{Intermediate embedding after 20 epochs}\label{fig:mammals_intermediate}
  \end{minipage}
  \hspace{2em}
  \begin{minipage}[b]{0.45\linewidth}
    \includegraphics[width=\linewidth]{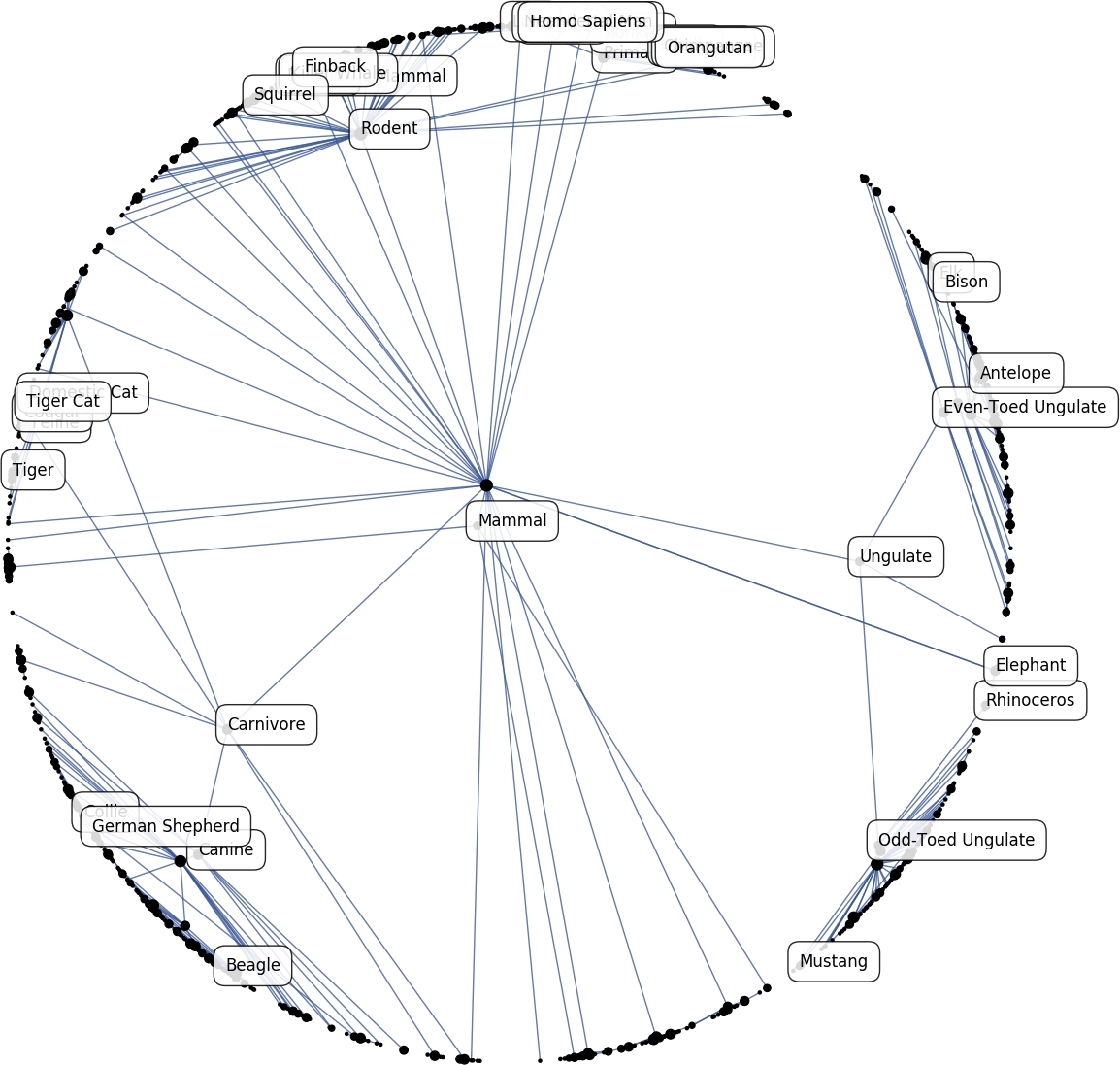}
    \subcaption{Embedding after convergence}\label{fig:mammals_converged}
  \end{minipage}
  \caption{Two-dimensional Poincaré embeddings of transitive closure of the \method{WordNet} mammals subtree. Ground-truth \texttt{is-a} relations of the original \method{WordNet} tree are indicated via blue edges.  
  A Poincaré embedding with $d=5$ achieves mean rank 1.26 and MAP 0.927 on this subtree. \label{fig:mammals-viz}}
\end{figure}

We evaluate the quality of the embeddings as commonly done for graph embeddings
\cite{DBLP:conf/nips/BordesUGWY13,kgs/nickel2016holographic}: For each observed
relationship \((u, v)\), we rank its distance \(d(\vu,\vv)\) among the ground-truth
negative examples for \(u\), i.e., among the set \({\{d(\vu, \vv^\prime)\ |\ (u,
v^\prime) \not \in \Set{D})\}}\). In the Reconstruction setting, we evaluate the
ranking on all nouns in the dataset. We then record the mean rank of \(v\) as well
as the mean average precision (MAP) of the ranking. The results of these
experiments are shown in \Cref{tab:embedding-nouns}. It can be seen that
Poincaré embeddings are very successful in the embedding of large taxonomies --
both with regard to their representation capacity and their generalization
performance. Even compared to Translational embeddings, which have more
information about the structure of the task, Poincaré embeddings show a greatly
improved performance while using an embedding that is smaller by an order of
magnitude. Furthermore, the results of Poincaré embeddings in the link
prediction task are very robust with regard to the embedding dimension.
We attribute this result to the structural bias of Poincaré embeddings, what
could lead to reduced overfitting on this kind of data with a clear latent
hierarchy. In \Cref{fig:mammals-viz} we show additionally a visualization of a
two-dimensional Poincaré embedding. For purpose of clarity, this embedding has
been trained only on the mammals subtree of \method{WordNet}.

\subsection{Network Embeddings}
\label{sec:orgeb02796}
Next, we evaluated the performance of Poincaré embeddings for link prediction in
networks. Since edges in complex networks can often be explained via latent
hierarchies over their nodes \cite{clauset2008hierarchical}, we are interested in the benefits of
Poincaré embeddings both in terms representation size and generalization
performance. We performed our experiments on four commonly used social networks,
i.e, \method{AstroPh}, \method{CondMat}, \method{GrQc}, and \method{HepPh}.
These networks represent scientific collaborations such that there exists an
undirected edge between two persons if they co-authored a paper. For these
networks, we model the probability of an edge as proposed by \citet{hyperbolic/krioukov2010hyperbolic}
via the Fermi-Dirac distribution
\begin{equation}
P((u,v) = 1\ |\ \Theta) = \frac{1}{e^{(d(\vu,\vv) - r)/ t} + 1} \label{eq:fermi}
\end{equation}
where \(r, t > 0\) are hyperparameters. Here, \(r\) corresponds to the
radius around each point \(\vu\) such that points within this radius are likely to
have an edge with \(u\). The parameter \(t\) specifies the steepness of the logistic
function and influences both average clustering as well as the degree
distribution \cite{hyperbolic/krioukov2010hyperbolic}. We use the
cross-entropy loss to learn the embeddings and sample negatives as in
\Cref{sec:taxonomies}.

For evaluation, we split each dataset randomly into train, validation, and test
set. The hyperparameters \(r\) and \(t\) where tuned for each method on the
validation set. \Cref{tab:network} lists the MAP score of Poincaré and Euclidean
embeddings on the test set for the hyperparameters with the best validation
score. Additionally, we again list the reconstruction performance without
missing data. Translational embeddings are not applicable to these datasets as
they consist of undirected edges. It can be seen that Poincaré embeddings
perform again very well on these datasets and -- especially in the
low-dimensional regime -- outperform Euclidean embeddings.

\begin{table}
  \small
  \centering
  \caption{Mean average precision for Reconstruction and Link Prediction on network data.
    \label{tab:network}}
  \begin{tabular}{llcccccccc}
    \toprule
    & & \multicolumn{8}{c}{\textbf{Dimensionality}}\\
    \cmidrule(l){3-10}
    & & \multicolumn{4}{c}{\textbf{Reconstruction}} & \multicolumn{4}{c}{\textbf{Link Prediction}}\\
    \cmidrule(lr){3-6} \cmidrule(l){7-10}
    & & 10 & 20 & 50 & 100 & 10 & 20 & 50 & 100 \\
    \midrule
    \method{AstroPh} & \textbf{Euclidean} & 0.376 & 0.788 & 0.969 & 0.989 
       & 0.508 & 0.815 & 0.946 & 0.960\\
    {\tiny N=18,772; E=198,110}& \textbf{Poincaré} & 0.703 & 0.897 & 0.982 & 0.990 
       & 0.671 & 0.860 & 0.977 & 0.988 \\
    \midrule
    \method{CondMat} & \textbf{Euclidean} & 0.356 & 0.860 & 0.991 & 0.998 
       & 0.308 & 0.617 & 0.725 & 0.736 \\
    {\tiny N=23,133; E=93,497} & \textbf{Poincaré} & 0.799 & 0.963 & 0.996 & 0.998 
       & 0.539 & 0.718 & 0.756 & 0.758 \\
    \midrule
    \method{GrQc} & \textbf{Euclidean} & 0.522 & 0.931 & 0.994 & 0.998 
       & 0.438 & 0.584 & 0.673 & 0.683 \\
    {\tiny N=5,242; E=14,496}& \textbf{Poincaré} & 0.990 & 0.999 & 0.999 & 0.999 
       & 0.660 & 0.691 & 0.695 & 0.697 \\
    \midrule
    \method{HepPh} & \textbf{Euclidean} & 0.434 & 0.742 & 0.937 & 0.966 
       & 0.642 & 0.749 & 0.779 & 0.783 \\
    {\tiny N=12,008; E=118,521} & \textbf{Poincaré} & 0.811 & 0.960 & 0.994 & 0.997 
       & 0.683 & 0.743 & 0.770 & 0.774 \\
    \bottomrule
  \end{tabular}
\end{table}

\subsection{Lexical Entailment}
\label{sec:org29598e6}
An interesting aspect of Poincaré embeddings is that they allow us to make
graded assertions about hierarchical relationships as hierarchies are
represented in a continuous space. We test this property on \method{HyperLex}
\cite{vulic2016hyperlex}, which is a gold standard resource for evaluating how
well semantic models capture graded lexical entailment by quantifying to what \emph{degree} \(X\) is a type of \(Y\) via
ratings on a scale of \([0,10]\). Using the noun part of \method{HyperLex}, which
consists of 2163 rated noun pairs, we then evaluated how well Poincaré
embeddings reflect these graded assertions. For this purpose, we used the
Poincaré embeddings that were obtained in \Cref{sec:taxonomies} by embedding
\method{WordNet} with a dimensionality \(d=5\). Note that these embeddings were
not specifically trained for this task. To determine to what
extent \(\texttt{is-a}(u, v)\) is true, we used the score function:
\begin{equation}
\score(\texttt{is-a}(u,v)) = -(1 + \alpha (\|\vv\| - \|\vu\|)) d(\vu, \vv) \label{eq:hyperlex-score} .
\end{equation}
Here, the term \(\alpha(\|\vv\| - \|\vu\|)\) acts as a penalty
when \(v\) is lower in the embedding hierarchy, i.e., when \(\vv\) has a higher norm
than \(\vu\). The hyperparameter \(\alpha\) determines the severity of the penalty. In
our experiments we set \(\alpha = 10^3\).

Using \Cref{eq:hyperlex-score}, we scored all noun pairs in \method{HyperLex}
and recorded Spearman's rank correlation with the ground-truth ranking. The
results of this experiment are shown in \Cref{tab:hyperlex-spearman}.
It can be seen that the ranking based on Poincaré embeddings clearly outperforms
all state-of-the-art methods evaluated in \cite{vulic2016hyperlex}. Methods in
\Cref{tab:hyperlex-spearman} that are prefixed with WN also use \method{WordNet}
as a basis and therefore are most comparable. The same embeddings also achieved
a state-of-the-art accuracy of \(0.86\) on \method{Wbless}
\cite{weeds2014learning,kiela2015exploiting}, which evaluates non-graded lexical
entailment.

\begin{table}[t]\small
\caption{\label{tab:hyperlex-spearman}
Spearman's \(\rho\) for Lexical Entailment on \method{HyperLex}.}
\centering
\begin{tabular}{lcccccccc}
\toprule
 & \textbf{FR} & \textbf{SLQS-Sim} & \textbf{WN-Basic} & \textbf{WN-WuP} & \textbf{WN-LCh} & \textbf{Vis-ID} & \textbf{Euclidean} & \textbf{Poincaré}\\
\midrule
\(\rho\) & 0.283 & 0.229 & 0.240 & 0.214 & 0.214 & 0.253 & 0.389 & 0.512\\
\bottomrule
\end{tabular}
\end{table}

\section{Discussion and Future Work}
\label{sec:org466597f}
In this paper, we introduced Poincaré embeddings for learning representations of
symbolic data and showed how they can simultaneously learn the
similarity and the hierarchy of objects. Furthermore, we proposed an efficient
algorithm to compute the embeddings and showed experimentally, that Poincaré
embeddings provide important advantages over Euclidean embeddings on
hierarchical data: First, Poincaré embeddings enable very parsimonious
representations whats allows us to learn high-quality embeddings of large-scale
taxonomies. Second, excellent link prediction results indicate that hyperbolic
geometry can introduce an important structural bias for the embedding of complex
symbolic data. Third, state-of-the-art results for predicting lexical entailment
suggest that the hierarchy in the embedding space corresponds well to the
underlying semantics of the data.

The main focus of this work was to evaluate the general properties of hyperbolic
geometry for the embedding of symbolic data. In future work, we intend, to both expand
the applications of Poincaré embeddings -- for instance to multi-relational data
-- and also to derive models that are tailored to specific applications such as
word embeddings. Furthermore, we have shown that natural gradient based
optimization already produces very good embeddings and scales to large
datasets. We expect that a full Riemannian optimization approach can further
increase the quality of the embeddings and lead to faster convergence.

\bibliographystyle{plainnat}
\bibliography{paper_biber}
\end{document}

%% file: geodesics.pgf
\pgfdeclarelayer{background}
\pgfdeclarelayer{foreground}
\pgfsetlayers{background,main,foreground}
\begin{tikzpicture}[very thick,scale=0.5]
\begin{pgfonlayer}{foreground}\draw (0,0) circle (3.0);\end{pgfonlayer}
\tikzstyle{segment}=[line width=0.3mm]

\begin{pgfonlayer}{background}
\end{pgfonlayer}

\draw[black] (-1.142, 2.774) arc (22.383:-22.383:7.285);
\draw[black] (-0.333, 2.981) arc (-173.619:-36.381:1.175);
\draw[black] (-1.500, -2.598) -- (1.500, 2.598);
\draw[segment,magenta] (0.000, 0.000) -- (1.142, 1.979);
\draw[segment,orange] (0.000, 2.285) arc (-135.230:-74.770:1.175);
\draw[segment,cyan] (-0.693, 1.201) arc (9.486:-9.486:7.285);

\begin{pgfonlayer}{foreground}
\draw[fill=black,black] (0.000,0.000) circle (0.05);
\draw[fill=black,black] (-0.693,-1.201) circle (0.05);
\draw[fill=black,black] (1.142,1.979) circle (0.05);
\draw[fill=black,black] (-0.693,1.201) circle (0.05);
\draw[fill=black,black] (0.000,2.285) circle (0.05);
\node at (0.4,-0.2) {\tiny $p_1$};
\node at (-0.3,-1.4) {\tiny $p_2$};
\node at (1.55,1.77) {\tiny $p_3$};
\node at (-1,1) {\tiny $p_4$};
\node at (-0.2,2) {\tiny $p_5$};
\end{pgfonlayer}
\end{tikzpicture}